# Modeling Clinical Concern Trajectories in Language Model Agents

Sukesh Subaharan[†]   VS Venkatesan[†]   P Murugadasan[†]   D Sivakumar[†]

N Gautham[†]   M Ganeshkumar[†]

[†]Coimbatore Medical College

Coimbatore, TN

India

email: sukeshsubaharan@yahoo.com / sukeshsubaharan05@gmail.com

## ABSTRACT

Large language model (LLM) agents deployed in clinical settings often exhibit abrupt, threshold-driven behavior, offering little visibility into accumulating risk prior to escalation. In real-world care, however, clinicians act on gradually rising concern rather than instantaneous triggers. We study whether explicit state dynamics can expose such pre-escalation signals without delegating clinical authority to the agent. We introduce a lightweight agent architecture in which a memoryless clinical risk encoder is integrated over time using first- and second-order dynamics to produce a continuous escalation pressure signal. Across synthetic ward scenarios, stateless agents exhibit sharp escalation cliffs, while second-order dynamics produce smooth, anticipatory concern trajectories despite similar escalation timing. These trajectories surface sustained unease prior to escalation, enabling human-in-the-loop monitoring and more informed intervention. Our results suggest that explicit state dynamics can make LLM agents more clinically legible by revealing *how long* concern has been rising, not just *when* thresholds are crossed.

# 1    Introduction

Large language model (LLM) agents are increasingly deployed in clinical decision-making, demonstrating promising performance in emergency triage [1,2], diagnostic reasoning [3,4], and multi-agent collaborative frameworks that produce structured explanations to increase clinician confidence [5]. Despite these advances, deployment in high-stakes clinical environments raises fundamental questions about interpretability, temporal reasoning, and the legibility of automated decision processes to human supervisors.

A critical limitation of current clinical AI systems lies in their threshold-driven behavior, which produces abrupt escalation decisions with limited visibility into accumulating risk. Continuous vital sign monitoring systems in surgical wards have revealed practical challenges: median alert acknowledgement times exceeding 100 minutes [6], alert fatigue, and mixed evidence for outcome improvements [7,8]. Ward-level surveillance systems analyzing millions of vital signs produce alerts at rates of 0.4 per clinician shift, with 88% of accurate alerts associated with downstream clinical actions [9], yet the operational challenge of surfacing rising risk, before thresholds are crossed, persists. Randomized trials of continuous wireless monitoring in postoperative patients have demonstrated implementation feasibility but inconclusive outcome effects [7,8,10], suggesting that threshold-based alerting is insufficient for capturing the temporal evolution of patient state and the gradual accumulation of physiological instability preceding frank deterioration.

Explainability alone does not address the temporal opacity of threshold-driven systems, which provide no indication of how long concern has been rising or whether instability represents transient noise or sustained deterioration. Argumentation-based LLM frameworks have shown that structured reasoning graphs can improve decision accuracy and user confidence [5], and explainable machine learning systems have improved interobserver agreement among clinicians [11]. However, decision-set approaches combining rule-based interpretability with reliability estimates [12] remain largely memoryless, lacking capacity to distinguish isolated abnormalities from progressive decline.

Temporal state estimation frameworks offer a solution by explicitly modeling patient state evolution. Deep state-space models applied to longitudinal electronic health record data have learned interpretable latent states associated with prognosis, improving temporal clustering and visualization of disease trajectories [13]. In sepsis management, hybrid approaches combining mechanistic cardiovascular physiology with distributional reinforcement learning have produced uncertainty-aware treatment policies that identify high-risk states consistent with clinical knowledge [14], while sequential decision frameworks for partially observed processes illustrate how explicit belief tracking enables optimal intervention policies [15]. Recent work in affective agent modeling has demonstrated that imposing explicit dynamical structure on agent-level state can induce temporal coherence and controlled recovery, with second-order momentum-based systems producing qualitatively distinct trajectories from stateless baselines [16]. These methods share a common principle: maintaining an internal representation of accumulated state enables smoother, more anticipatory behavior that mirrors human reasoning.

The application of explicit state dynamics to safety-critical clinical monitoring, where escalation decisions carry immediate patient safety implications and must remain legible to human supervisors, represents an unexplored frontier. We address this gap by adapting explicit state dynamics from affective agent modeling [16] to clinical risk assessment in longitudinal ward monitoring. We introduce a lightweight agent architecture in which a memoryless clinical risk encoder is integrated over time using first- and second-order dynamics to produce a continuous escalation pressure signal. By incorporating asymmetric smoothing and velocity-based hysteresis, second-order dynamics model the clinical principle that worsening signals propagate faster than apparent improvements, producing concern trajectories that rise more consistently under sustained deterioration. Across synthetic ward scenarios capturing common postoperative deterioration patterns, we demonstrate that explicit state dynamics transform sharp escalation cliffs into smooth, legible trajectories revealing how long concern has been rising, not just when thresholds are crossed.

Our contributions are threefold. First, we demonstrate that explicit temporal dynamics, previously validated in affective agent modeling [18], can be adapted to clinical risk assessment to produce interpretable escalation behavior. Second, we introduce metrics, unease lead time (ULT), unease area (UA), and escalation jerk (EJ), that quantify temporal characteristics of escalation pressure trajectories, capturing anticipatory signaling, cumulative hesitation, and smoothness critical for clinical legibility. Third, we provide empirical evidence that second-order dynamics with hysteresis produce escalation trajectories more clinically legible to human supervisors than stateless or first-order alternatives, despite similar escalation timing. Our results suggest that explicit state dynamics offer a path toward making LLM agents more transparent and trustworthy in longitudinal clinical monitoring by revealing the temporal evolution of concern rather than presenting escalation as an instantaneous, threshold-driven event.

## 2 Methods

### 2.1 Task Definition

We study escalation behavior in longitudinal surgical ward monitoring, where an agent must decide when to escalate care as patient physiology evolves over time.

Each episode consists of a discrete-time sequence

$$\{x_t\}_{t=1}^T$$

where $x_t$ is a structured snapshot of the patient state at hour $t$, including vital signs, mental status and urine output trends. At each timestep, the agent produces an internal continuous risk representation,

$$z_t \in [0, 1]^3$$

a scalar escalation pressure,

$$E_t \in [0,1]$$

and a discrete action

$$a_t \in \{A_0, A_1, A_2\}$$

corresponding to conservative management, heightened vigilance or escalation. The underlying language model is frozen and used only to generate clinical notes consistent with the selected action.

### 2.2 Physiological Invariants and Violation Indicators

Each snapshot $x_t$ is mapped to a set of binary invariant violations

$$v_t \in \{0, 1\}^K$$

encoding deviations from physiological safety bounds [17-20]: hypotension (mean arterial pressure (MAP) <75mmHg, or <70mmHg, for severe reduction), hypoxemia (SpO$_2$ < 94%, or <92% for severe hypoxemia), tachypnea (respiratory rate (RR) > 22), fever (temperature ≥ 38.3 degrees Celsius), abnormal mental status and reduced urine output. These invariants are not learned and reflect widely accepted ward-level safety thresholds.

### 2.3 Instantaneous Risk Encoding

Each snapshot is first mapped to an instantaneous risk vector

$$r_t = (S_t, U_t, C_t) \in [0, 1]^3$$

representing numerical indicators of stability (S), urgency (U) and control margin (C) at given time $t$. The encoder is deterministic and piecewise linear, with the following mathematical representations. Let 1[.] denote the indicator function, equal to 1 when its argument is true and 0 otherwise. Stability at time $t$ captures the degree to which physiological parameters remain within acceptable ranges,

$$S_t = max(0, 1 - 0.3\ 1[MAP_t < 75] - 0.3\ 1[SpO2_t < 94] - 0.2\ 1[RR_t > 22] - 0.2\ 1[Temp_t \geq 38.3])$$

The weights reflect relative clinical salience while ensuring $S_t \in [0,1]$.

Urgency at time $t$ represents the immediacy with which intervention may be required, independent of trend of history. It is defined as,

$$U_t = min(1, 0.4\ 1[MAP_t < 75] + 0.3\ 1[SpO2_t < 94] + 0.2\ 1[RR_t > 22] + 0.1\ 1[Temp_t \geq 38.3])$$

Unlike stability, urgency accumulates additively with concurrent abnormalities, reflecting compounding clinical risk.

Control margin captures loss of physiological or neurological reserve that may impair safe observation. It is defined as

$$C_t = 1 - 0.4\,\mathbb{1}[mental\ status\ abnormal] - 0.4\,\mathbb{1}[urine\ output\ reduced]$$

This dimension encodes deterioration that may not immediately manifest as vital sign instability but reduces tolerance for delayed escalation.

The encoder's intentionally memoryless, bounded, and non-learned nature ensures that any observed difference in escalation behavior across agents arise from the temporal integration and hysteresis mechanisms, not from differences in feature extraction or implicit state accumulation. The weights used in the instantaneous risk encoder are heuristic and chosen to reflect coarse relative salience of common ward-level abnormalities rather than to model disease-specific risk. Larger penalties are assigned to hypotension and hypoxemia due to their broad association with acute instability, while tachypnea and fever contribute smaller penalties. Control margin emphasizes neurological status and urine output as indicators of reduced physiological reserve. Weights are fixed across all experiments and are not tuned to optimize any outcome.

### 2.4 Latent risk dynamics

We compare three dynamical systems that transform instantaneous risk $r_t$ into a latent accumulated state $z_t$.

#### 2.4.1 Stateless baseline

This corresponds to a purely reactive agent with no temporal memory.

$$z_t = r_t$$

#### 2.4.2 First-order dynamics: Exponential smoothing

$$z_t = \begin{cases} r_t, & t = 1 \\ \alpha r_t + (1-\alpha)z_{t-1}, & t > 1 \end{cases}$$

With $\alpha = 0.3$. This implements leaky integration of risk, equivalent to a discrete-time first-order low-pass filter.

#### 2.4.3 Second-order hysteretic dynamics

To model escalation inertia and resistance to abrupt de-escalation, we implement an asymmetric second-order system with velocity. Direction dependent smoothing is modeled as

$$\hat{z}_t = \alpha_t r_t + (1-\alpha_t)z_{t-1}$$

where

$$\alpha_t = \begin{cases} \alpha_{\text{down}}, & \exists i: r_{t,i} < z_{t-1,i} \\ \alpha_{\text{up}}, & \text{otherwise} \end{cases}$$

with $\alpha_{\text{down}} = 0.4$, $\alpha_{\text{up}} = 0.15$.

Velocity of updates are modelled as

$$\boldsymbol{v}_t = \beta(\hat{\boldsymbol{z}}_t - \boldsymbol{z}_{t-1}) + (1-\beta)\boldsymbol{v}_{t-1}$$

With $\beta = 0.6$.

The state update is carried out as

$$\boldsymbol{z}_t = clip(\hat{\boldsymbol{z}}_t + \boldsymbol{v}_t, 0, 1)$$

The setup induces hysteresis by propagation of worsening physiological signal faster than the apparent improvement. Asymmetric smoothing coefficients ($\alpha_{\text{down}} > \alpha_{\text{up}}$) are used to model escalation hysteresis: worsening physiology is incorporated more rapidly than apparent improvement. This asymmetry reflects a conservative clinical bias against premature de-escalation under noisy measurements. The velocity term introduces inertia, discouraging abrupt reversals in latent state. Parameter values were selected to induce gentle hysteresis rather than discrete state changes and were not optimized.

## 2.5 Escalation pressure and action selection

The latent state $\boldsymbol{z}_t = (S_t, U_t, C_t)$ is mapped to a scalar escalation pressure, defined as a weighted linear combination of instability $(1-S)$, urgency $(U)$, and loss of control margin $(1-C)$. Equal weight is assigned to instability and urgency, reflecting their primary role in acute deterioration, while control margin contributes a smaller but non-negligible component. The resulting formulation balances sensitivity to immediate physiological abnormalities with contextual loss of reserve, producing a smooth scalar signal rather than a discrete decision rule.

$$E_t = 0.4(1 - S_t) + 0.4 U_t + 0.2(1 - C_t)$$

The action is selected via fixed thresholds,

$$a_t = \begin{cases} A_0, & E_t < 0.4 \\ A_1, & 0.4 < E_t < 0.6 \\ A_2, & E_t \geq 0.6 \end{cases}$$

Where only $A_2$ is considered an explicit escalation.

Action thresholds were selected to partition escalation pressure into low ($E < 0.4$), intermediate ($0.4 \leq E < 0.6$), and high ($E \geq 0.6$) concern regimes. The escalation threshold ($E \geq 0.6$) was chosen to correspond to a clinically conservative operating point: escalation occurs only when multiple dimensions of instability are concurrently present or when deterioration is sustained over time. This choice reflects common ward practice, where escalation is typically triggered by combinations of abnormalities rather than isolated deviations. The threshold is fixed by design and not treated as an emergent property of the model. This formulation is intended to produce a smooth, interpretable scalar signal rather than a calibrated risk score. All weights are fixed and shared across agents to isolate the effect of temporal dynamics.

## 2.6 Evaluation metrics

Let τ denote the agent's first escalation time.

We report escalation status as three possible outcomes, early escalation (τ < $t_{min}$), late escalation (τ > $t_{max}$) or escalation within the acceptable window. Unease Lead Time (ULT) is a metric defined to represent the delay between the first elevated $E_t > 0.3$ and escalation. Unease Area (UA) was defined to quantify the cumulative exposure to escalation pressure prior to escalation, reflecting prolonged hesitation under increasing risk. It is defined as

$$UA = \begin{cases} \sum_{t<\tau} E_t, & \text{if escalation occurs} \\ \sum_{t=1}^{T} E_t, & \text{otherwise} \end{cases}$$

Additionally, Escalation Jerk (EJ) is defined to capture the maximum abrupt change in escalation pressure, quantifying temporal smoothness and resistance to sudden reactive escalation. It is represented as

$$EJ = \max_{t} |E_t - E_{t-1}|$$

## 2.7 Synthetic Trajectory Generation and Noise Injection

To evaluate escalation dynamics under controlled yet diverse longitudinal conditions, we use synthetically generated patient trajectories.

Initial base trajectories were generated using a large instruction-tuned language model (DeepSeek), prompted to produce hour-by-hour ward-level physiological snapshots consistent with common postoperative and acute surgical deterioration patterns (prompt used detailed in Supplementary Appendix S1) [21]. These trajectories specify vital signs, mental status, and urine output trends at each timestep. Synthetic generation was chosen to allow precise control over temporal progression (slow drift vs abrupt deterioration), repeatable evaluation across agent variants, and stress-testing of escalation dynamics under rare but clinically plausible patterns.

*2.7.1 Stochastic Perturbation of Physiological Signals*

To reduce unrealistically smooth trajectories and introduce measurement irregularities commonly observed in real ward data, we applied lightweight stochastic perturbations to the generated snapshots. Specifically, for a subset of timesteps, individual physiological variables were subjected to small random deviations, including transient abrupt changes (single-step spikes or drops), short-lived reversals followed by reversion to trend, isolated noisy readings inconsistent with adjacent timesteps. These perturbations were applied sparingly and independently across variables, preserving overall clinical plausibility while breaking monotonicity assumptions. Importantly, noise was injected only after base trajectory generation. Perturbations did not systematically bias escalation timing. Ground-truth escalation windows were computed after noise injection, ensuring consistent evaluation.

This process is intended to approximate real-world charting artifacts (measurement noise, transient instability, delayed documentation) rather than adversarial corruption. We emphasize that the purpose of stochastic perturbation is not realism at the level of physiological simulators, but to ensure that escalation behavior is not trivially driven by smooth or perfectly monotonic inputs. All agent variants are evaluated on identical perturbed trajectories, isolating the effect of latent dynamics and temporal integration, rather than sensitivity to clean inputs.

*2.8  Language Model Integration*

A frozen instruction-tuned 7B large language model (LLM) is used solely to generate narrative plans conditioned on the chosen action and latent state [22]. The prompt exposes the internal state explicitly (stability, urgency, control margin) and constrains the model to produce text consistent with the selected action (Prompt detailed in Supplementary Appendix S2).

# 3  Results

*3.1  Temporal Escalation Trajectories*

We examine the escalation pressure trajectories $E_t$ to qualitatively assess temporal behavior across agent variants. Figure 1 shows representative runs for each agent under abrupt, slow-drift and stable trajectories. The stateless agent exhibits reactive dynamics, with escalation pressure remaining low until abrupt jumps triggered by instantaneous abnormalities. In contrast, the first-order agent demonstrates gradual accumulation of escalation pressure under sustained deterioration, but frequently attenuates concern following transient improvement.

The second-order hysteretic agent exhibits smooth, monotonic increases in escalation pressure during worsening physiology and sustained elevation despite short-lived reversals, reflecting resistance to premature de-escalation. This behavior is particularly evident in trajectories containing stochastic perturbations.

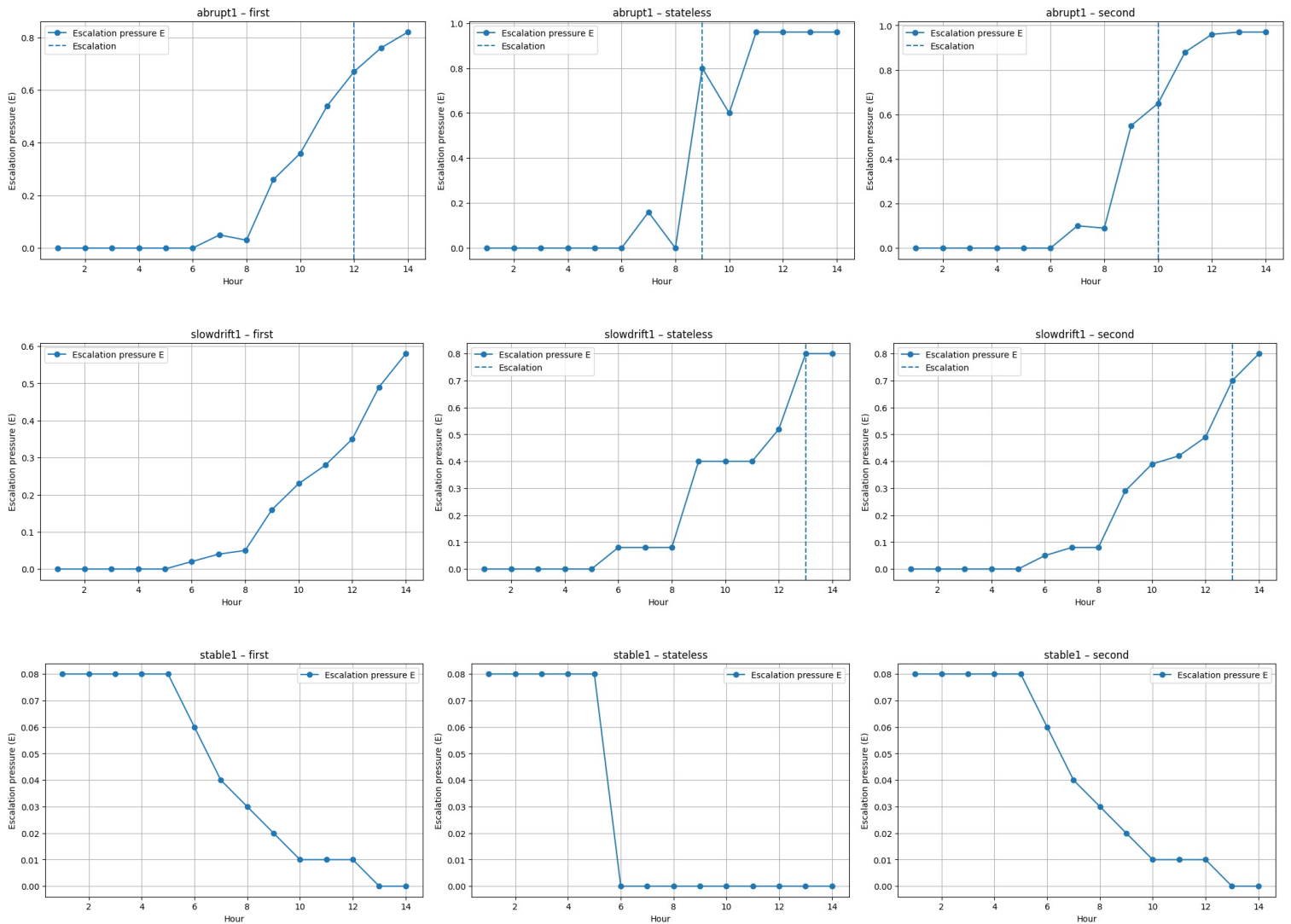

**Figure 1:** Representative escalation pressure trajectories ($E_t$) for stateless, first-order, and second-order agents under noisy longitudinal deterioration (abrupt and slow deterioration) and stable. While instantaneous differences are modest, second-order hysteretic dynamics exhibit smoother accumulation and reduced backtracking of escalation pressure, visible as fewer sharp inflections when trajectories are viewed at scale. Trajectories are shown for interpretability rather than pointwise comparison.

### 3.2   Escalation Timing and Unease Lead Time

We quantify anticipatory behavior using unease lead time (ULT), defined as the delay between initial elevation of escalation pressure and escalation action. Figure 2 summarizes ULT across agents. The stateless agent exhibits the longest mean ULT with high variance, reflecting delayed escalation following prolonged low concern. The first-order agent reduces ULT, while the second-order agent achieves intermediate ULT with reduced variability.

This pattern indicates that hysteretic dynamics promote balanced recognition of deterioration without excessively prolonging escalation delay.

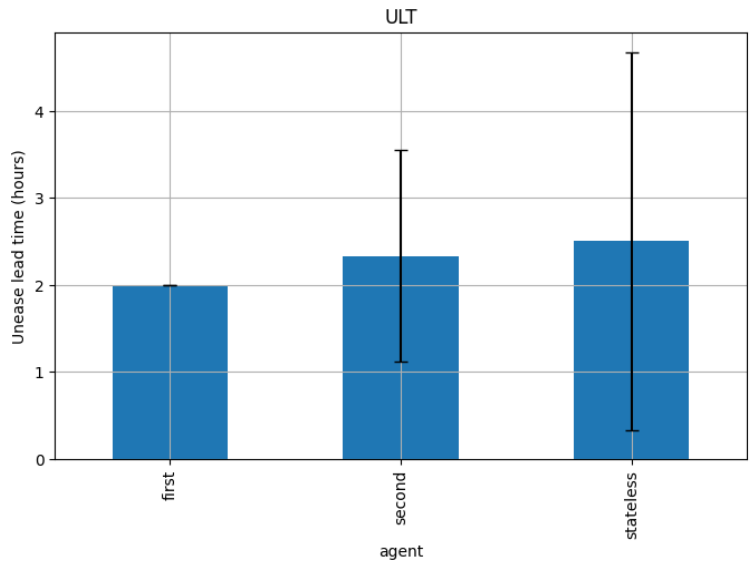

**Figure 2:** Unease Lead Time (ULT), measuring the delay between initial elevation of escalation pressure and escalation. Differences are shown for completeness; escalation timing itself is not a primary outcome of this study. Error bars denote mean ± standard deviation.

## 3.3 Summary Across Metrics

Table 1 summarizes mean and standard deviation of ULT, UA, and EJ across all evaluated trajectories. Across all metrics, second-order hysteretic dynamics strike a consistent balance: reduced escalation jerk, moderate unease accumulation, and anticipatory escalation timing. These improvements arise from explicit temporal dynamics rather than changes to the underlying language model.

| Agent Variant | ULT | | UA | | EJ | |
|---|---|---|---|---|---|---|
| | mean | std | mean | std | mean | std |
| First-order | 2.000 | 0.000 | 1.314 | 0.721 | 0.126 | 0.091 |
| Second-order | 2.333 | 1.211 | 1.040 | 0.717 | 0.223 | 0.187 |
| Stateless | 2.500 | 2.168 | 0.924 | 0.930 | 0.382 | 0.301 |

**Table 1:** Summary of evaluation metrics ULT, UA, EJ reported with standard deviation

### 3.4 Accumulated Escalation Pressure (Unease Area)

We next evaluate Unease Area (UA), which captures cumulative escalation pressure prior to escalation. As shown in Figure 3, stateless agents exhibit the lowest mean UA, but with substantial variance, reflecting either premature escalation or prolonged inaction. First-order agents show the highest UA, consistent with sustained concern that dissipates slowly before escalation. Second-order agents reduce UA relative to first-order dynamics, indicating that escalation pressure accumulates more consistently but does not persist excessively before action. Together with ULT, these results suggest that hysteresis enables relatively early yet more decisive escalation.

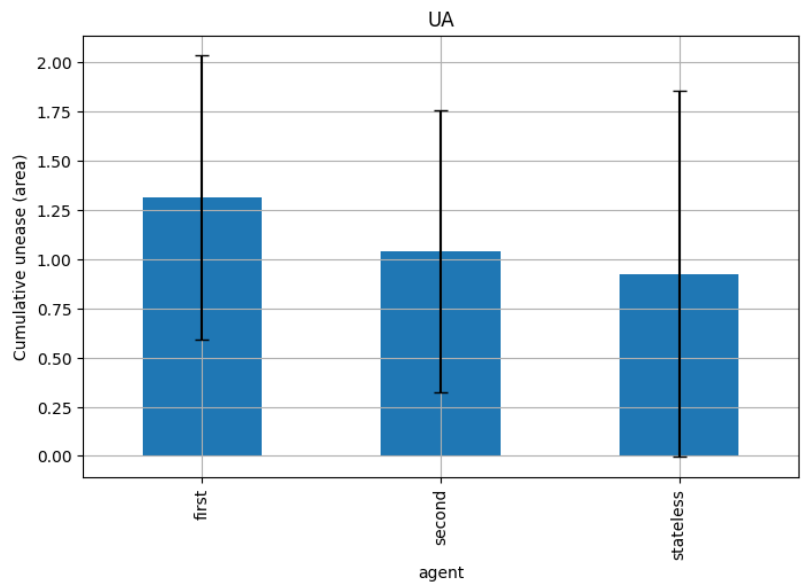

**Figure 3:** Unease Area (UA, cumulative escalation pressure prior to escalation or over the full trajectory) across agent variants. First-order dynamics exhibit the largest accumulated unease, consistent with sustained concern that attenuates slowly. Second-order dynamics reduce cumulative unease relative to first-order integration while maintaining smooth accumulation. Error bars denote mean ± standard deviation.

### 3.5 Temporal Smoothness and Escalation Jerk

To assess abruptness of escalation behavior, we measure Escalation Jerk (EJ), defined as the maximum absolute change in escalation pressure between consecutive timesteps. As shown in Figure 4, stateless agents exhibit the highest EJ, consistent with reactive escalation driven by instantaneous changes in input. Introducing temporal integration substantially reduces jerk: first-order dynamics achieve the lowest mean EJ, while second-order hysteretic dynamics yield intermediate values.

Notably, second-order dynamics display greater variability in EJ than first-order smoothing, reflecting the presence of inertia and asymmetric update rules that allow rapid incorporation of worsening signals while resisting de-escalation. Overall, both stateful variants reduce abrupt escalation relative to the stateless baseline, with second-order dynamics trading minimal jerk for increased temporal persistence.

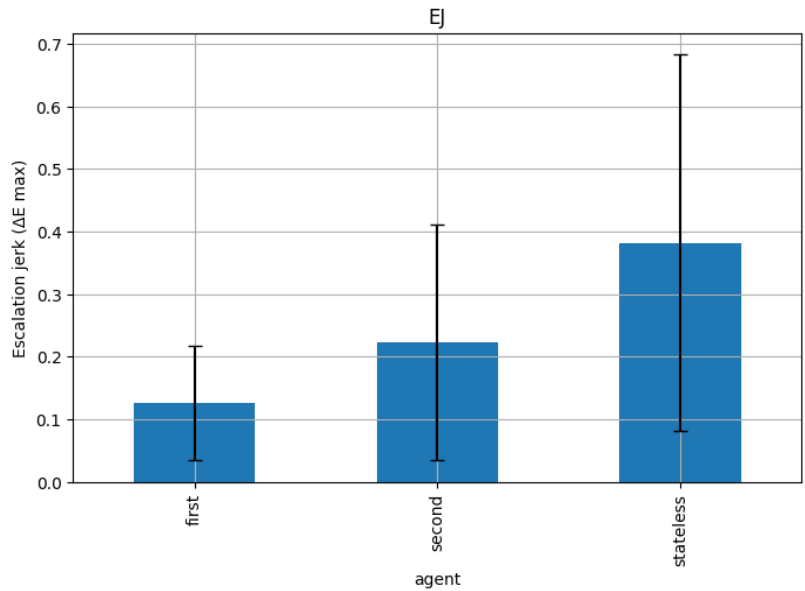

**Figure 4:** Escalation Jerk (EJ), defined as the maximum absolute change in escalation pressure between consecutive timesteps, across agent variants. Stateless agents exhibit the highest jerk, indicating abrupt reactive changes. Both stateful agents reduce jerk relative to the stateless baseline: first-order dynamics yield the lowest mean EJ, while second-order hysteretic dynamics show intermediate jerk with higher variability, consistent with inertial and asymmetric temporal updates. Error bars denote mean ± standard deviation across trajectories.

## 4 Discussion

### 4.1 Principal Findings

This study demonstrates that explicit state dynamics can make LLM agents more clinically legible in longitudinal ward monitoring by exposing pre-escalation signals that reveal how long concern has been rising, not just when thresholds are crossed. Across synthetic postoperative deterioration scenarios, second-order hysteretic dynamics exhibited favorable properties compared to stateless and first-order alternatives on metrics capturing temporal smoothness, anticipatory behavior, and cumulative risk exposure. These improvements arose entirely from temporal integration architecture rather than modifications to the underlying language model, suggesting that explicit state dynamics offer a generalizable approach to controlling agent behavior in safety-critical domains. Throughout this discussion, anticipatory behavior refers to earlier and smoother accumulation of escalation pressure under sustained deterioration, rather than correctness of escalation timing.

The results reveal a clear hierarchy in escalation behavior. Stateless agents exhibited the highest escalation jerk (mean 0.382), reflecting abrupt threshold-driven decisions with no temporal memory. First-order exponential smoothing reduced jerk substantially (mean 0.126) by introducing temporal continuity, but produced rigid trajectories with zero variance in unease lead time. Second-order hysteretic dynamics achieved intermediate jerk (mean 0.223) while introducing controlled variability in escalation timing (ULT std 1.211), indicating adaptive responses to different deterioration patterns rather than mechanical smoothing. This pattern aligns with clinical reasoning, where concern accumulates gradually but escalation timing depends on trajectory shape, not just instantaneous severity.

The unease area metric revealed that second-order dynamics achieved moderate accumulation (mean 1.040) compared to first-order systems (mean 1.314). This suggests that hysteretic integration surfaces concern in a balanced fashion without excessive delay, consistent with the clinical principle that sustained physiological instability warrants attention even before discrete thresholds are breached. The reduced standard deviation in unease area for second-order systems (0.717 vs 0.930 for stateless) indicates more consistent pre-escalation signaling across diverse trajectories, a property critical for human-in-the-loop monitoring where unpredictable alert behavior contributes to alarm fatigue [6].

### 4.2 Comparison with Existing Clinical Monitoring Systems

Current continuous vital sign monitoring systems deployed in surgical wards face operational challenges that our findings directly address. Implementations report median alert acknowledgement times exceeding 100 minutes [6] and mixed evidence for outcome improvements despite technological sophistication [7,8]. These challenges stem partly from threshold-driven alerting that provides no visibility into accumulating risk, forcing clinicians to react to discrete events rather than monitor evolving concern. Our second-order dynamics produce smooth escalation pressure trajectories that could enable proactive monitoring dashboards, where rising concern is visible hours before escalation thresholds are crossed.

Ward-level surveillance systems analyzing millions of vital signs produce alerts at rates of 0.4 per clinician shift, with 88% of accurate alerts associated with downstream clinical actions [9]. However, these systems remain reactive, by alerting only when thresholds are violated. By exposing unease lead time and cumulative risk exposure, explicit state dynamics could complement threshold-based systems by providing early warning signals that distinguish transient instability from sustained deterioration. This distinction is critical in ward settings where clinicians manage multiple patients and must prioritize attention efficiently.

Recent randomized trials of continuous wireless monitoring have demonstrated implementation feasibility but inconclusive outcome effects [7,8,10], suggesting that sensor technology alone is insufficient. Our results indicate that the temporal integration architecture, which refer to how instantaneous measurements are transformed into actionable signals, may be as important as measurement frequency or accuracy. Second-order dynamics with hysteresis model the clinical principle that worsening propagates faster than apparent improvement, producing trajectories that mirror how experienced clinicians accumulate concern over time.

*4.3    Implications for LLM Agent Design in Clinical Settings*

The growing deployment of LLM agents in clinical decision-making [1-5] raises fundamental questions about temporal reasoning and interpretability. Existing frameworks emphasize turn-local reasoning [3,4] or structured argumentation [5] but lack explicit mechanisms for temporal coherence across extended monitoring periods. Our adaptation of affective state dynamics [16] to clinical risk assessment demonstrates that domain-general architectural principles can transfer to safety-critical applications when properly grounded in domain constraints.

The separation of instantaneous risk encoding from temporal integration offers a principled approach to agent design. By maintaining a frozen, memoryless encoder and varying only the integration dynamics, we isolate the effect of temporal structure on escalation behavior. This modularity enables direct comparison of dynamical systems and facilitates clinical validation, as the encoder's physiological invariants remain interpretable and auditable regardless of integration method. In contrast, end-to-end learned systems that implicitly accumulate state in neural network weights offer less transparency about how temporal information influences decisions.

Explainability in clinical AI has focused primarily on feature attribution and reasoning transparency [5,11,12], but temporal explainability remains underexplored. Our unease lead time and unease area metrics provide quantitative handles on temporal behavior that could be surfaced in clinical interfaces. For example, a dashboard could display not just current escalation pressure but also the duration and cumulative area of elevated concern, enabling clinicians to distinguish between newly emerging instability and prolonged deterioration that has been building over hours.

*4.4    Relationship to Temporal State Estimation in Healthcare*

Our findings align with broader evidence that explicit temporal dynamics improve clinical decision support. Deep state-space models applied to longitudinal EHR data have demonstrated that learned latent states can capture disease progression and improve prognostic clustering [13]. In sepsis management, hybrid approaches combining mechanistic physiology with reinforcement learning have produced uncertainty-aware treatment policies that identify high-risk states consistent with clinical knowledge [14]. Sequential decision frameworks for partially observed processes illustrate how explicit belief tracking enables optimal intervention timing under uncertainty [15].

Our contribution extends this line of work by demonstrating that relatively simple dynamical systems such as exponential smoothing and momentum-based integration, can produce clinically relevant temporal behavior without requiring large-scale learning or complex mechanistic models. The second-order dynamics we employ are computationally lightweight, requiring only maintenance of a state vector and velocity term, making them practical for real-time ward monitoring where computational resources may be limited and interpretability is paramount.

The hysteresis introduced by asymmetric smoothing (faster upward integration, slower downward decay) reflects a fundamental principle in clinical monitoring: it is safer to maintain concern during apparent improvement than to rapidly dismiss accumulating evidence of deterioration. This asymmetry is difficult to encode in purely threshold-based systems but emerges naturally from second-order dynamics with directionally dependent smoothing. The resulting trajectories exhibit controlled recovery rather than abrupt de-escalation, mirroring how clinicians maintain vigilance even after transient improvements in vital signs.

### 4.5 Limitations and Future Directions

Several limitations warrant discussion. First, our evaluation used synthetically generated trajectories rather than real patient data. While synthetic scenarios enabled controlled comparison of dynamical systems under identical inputs, they cannot capture the full complexity of real ward environments, including measurement noise, documentation delays, concurrent interventions, and the heterogeneity of postoperative patient populations. Validation using retrospective clinical data with expert-annotated escalation windows is a critical next step.

Second, our physiological invariants and risk encoder weights were manually specified based on clinical guidelines [17-20] rather than learned from data. While this approach ensures interpretability and clinical grounding, it may not capture institution-specific practices or population-specific risk profiles. Hybrid approaches that learn encoder weights while maintaining explicit dynamics could offer a path toward personalization while preserving temporal legibility.

Third, we evaluated escalation behavior in isolation, without modeling the downstream effects of earlier or smoother escalation on clinical workflows, resource utilization, or patient outcomes. The clinical value of anticipatory signaling depends on whether earlier visibility into rising concern enables meaningful interventions. Prospective simulation studies that model clinician response to different escalation pressure trajectories could quantify the operational impact of explicit state dynamics.

Fourth, our fixed threshold-based action selection (escalate when pressure $\geq 0.6$) does not account for context-dependent escalation criteria, resource availability, or clinician judgment. Real clinical escalation involves multifactorial decision-making that considers patient goals, code status, and available interventions. Future work should explore how explicit state dynamics interact with more sophisticated action selection policies, including those that incorporate uncertainty quantification or multi-objective optimization.

Fifth, the language model in our architecture served only to generate narrative plans conditioned on selected actions, not to participate in risk assessment or escalation decisions. While this design preserves interpretability and clinical control, it does not leverage the reasoning capabilities that make LLM agents promising for clinical decision support [1-5]. Hybrid architectures that use explicit dynamics to constrain LLM behavior while allowing language-based reasoning about complex clinical scenarios represent an important frontier.

*4.6     Clinical Translation Considerations*

Translating explicit state dynamics into clinical practice requires addressing several implementation challenges. First, clinical interfaces must surface temporal information in ways that support rather than burden clinical workflows. Dashboards displaying escalation pressure trajectories, unease lead time, and cumulative risk exposure could provide situational awareness, but their design must account for cognitive load, alarm fatigue, and integration with existing monitoring systems.

Second, the parameterization of dynamical systems (smoothing constants, momentum coefficients, threshold values) must be validated and potentially tuned for specific clinical contexts. While our parameters were chosen to demonstrate qualitative differences between dynamical regimes, clinical deployment would require systematic optimization using historical data and prospective evaluation of false positive and false negative rates.

Third, clinician trust in automated monitoring systems depends on transparency about how decisions are made. The interpretability of our approach offers a foundation for clinical validation, but formal usability studies are needed to assess whether clinicians find explicit state dynamics more trustworthy than threshold-based alternatives.

Fourth, regulatory pathways for AI-based clinical decision support systems increasingly emphasize transparency, validation, and ongoing monitoring. The modular architecture we propose may facilitate regulatory review compared to end-to-end learned systems where temporal reasoning is implicit in neural network weights.

*4.7     Broader Implications for Agent-Based Clinical Decision Support*

Our results suggest that explicit state dynamics offer a generalizable approach to controlling LLM agent behavior in domains where temporal coherence and anticipatory signaling are critical. Beyond ward monitoring, applications could include outpatient chronic disease management (where concern accumulates over weeks), intensive care unit sedation protocols (where state transitions must be smooth to avoid physiological instability), and emergency department triage (where evolving patient state influences resource allocation).

The principle of separating instantaneous assessment from temporal integration applies broadly to sequential decision-making under uncertainty. In domains where human supervisors must maintain situational awareness and intervene when necessary, exposing the temporal evolution of agent state enables more effective human-in-the-loop collaboration. This approach preserves human authority while leveraging automated systems for continuous monitoring and early warning.

As LLM agents become more prevalent in clinical settings, architectural choices that promote temporal legibility and controlled behavior will be essential for safe deployment. Our findings demonstrate that relatively simple dynamical systems can produce qualitatively different agent behavior without modifying the underlying language model, offering a lightweight and interpretable approach to temporal reasoning in safety-critical domains.

## 5     Conclusion

Explicit state dynamics transform LLM agent escalation behavior from abrupt, threshold-driven decisions to smooth, anticipatory trajectories that reveal the temporal evolution of clinical concern. Second-order hysteretic dynamics achieve a favorable balance across metrics capturing smoothness, anticipatory timing, and cumulative risk exposure, suggesting that momentum-based integration with directionally dependent smoothing aligns with clinical reasoning principles. These improvements arise from temporal integration architecture rather than language model modifications, offering a modular and interpretable approach to agent design in safety-critical domains. These findings concern the temporal legibility of escalation behavior rather than the clinical correctness of escalation decisions. Future work should validate these findings using real patient data, explore hybrid architectures that combine explicit dynamics with LLM reasoning capabilities, and assess the clinical impact of anticipatory escalation signaling on workflow efficiency and patient outcomes.